\documentclass[conference]{IEEEtran}
\IEEEoverridecommandlockouts
\usepackage[left=1.62cm,right=1.62cm,top=1.9cm]{geometry}

\usepackage{times}
\usepackage{soul}
\usepackage{url}
\usepackage[hidelinks]{hyperref}
\usepackage[utf8]{inputenc}
\usepackage[small]{caption}
\usepackage{graphicx}
\usepackage{amsmath}
\usepackage{amsthm}
\usepackage{booktabs}
\usepackage[switch]{lineno}
\usepackage{url}
\usepackage{mathtools}
\usepackage{amssymb}
\usepackage{soul}

\usepackage{xcolor}
\usepackage{soul}
\usepackage{titlesec}
\usepackage[acronym]{glossaries}
\usepackage{comment}
\usepackage[normalem]{ulem}

\usepackage{caption}

\usepackage{soul}
\usepackage[linesnumbered,ruled]{algorithm2e}
\usepackage{titlesec}
\usepackage[acronym]{glossaries}
\usepackage{comment}
\titlespacing{\section}{0pt}{\parskip}{-\parskip}

\newacronym{ros}{ROS}{robot operating system}
\newacronym{imu}{IMU}{inertial measurement unit}
\newacronym{vr}{VR}{virtual reality }
\newacronym{ai}{AI}{artificial intelligence}
\newacronym{slam}{SLAM}{simultaneous localization and mapping}
\newacronym{ogm}{OGM}{occupancy grid map}
\newacronym{tcp}{TCP}{transmission control protocol}
\newacronym{ip}{IP}{internet protocol}
\newacronym{amcl}{AMCL}{adaptive monte carlo localization}
\newacronym{os}{OS}{operating system}
\newacronym{tf}{TF}{transform frames}
\newacronym{rviz}{RViz}{ROS visualization}
\newacronym{xr}{XR}{extended reality}
\newacronym{sdk}{SDK}{software development kit}
\newacronym{lidar}{LiDAR}{light detection and ranging}

\newcommand{\server}{S}
\newcommand{\A}{\alpha}
\newcommand{\B}{\beta}

\begin{document}

\title{Real-Time Remote Control via VR over Limited Wireless Connectivity
}
\author{
\IEEEauthorblockN{
H.P.~Madushanka\IEEEauthorrefmark{1},~\IEEEmembership{Student Member,~IEEE,}
Rafaela~Scaciota\IEEEauthorrefmark{1}\IEEEauthorrefmark{2},~\IEEEmembership{Member,~IEEE,}\\
Sumudu~Samarakoon\IEEEauthorrefmark{1}\IEEEauthorrefmark{2},~\IEEEmembership{Member,~IEEE}, and
Mehdi Bennis\IEEEauthorrefmark{1}~\IEEEmembership{Fellow,~IEEE}
}
\IEEEauthorblockA{
	\small%
	\IEEEauthorrefmark{1}%
	Centre for Wireless Communication, University of Oulu, Finland \\
	\IEEEauthorrefmark{2}%
 Infortech Oulu, University of Oulu, Finland \\
 email: \{madushanka.hewapathiranage, rafaela.scaciotatimoesdasilva, sumudu.samarakoon, mehdi.bennis\}@oulu.fi
}
\vspace{-20pt}
\thanks{
This work was supported by NSF-AKA CRUISE (GA 24304406), VERGE (GA 101096034), and Infotech-R2D2. Views and opinions expressed are however those of the author(s) only and do not necessarily reflect those of the European Union. Neither the European Union nor the granting authority can be held responsible for them.} 
}

\maketitle


\begin{abstract}
This work introduces a solution to enhance human-robot interaction over limited wireless connectivity. 
The goal is to enable remote control of a robot through a \gls{vr} interface, ensuring a smooth transition to autonomous mode in the event of connectivity loss. 
The \gls{vr} interface provides access to a dynamic 3D virtual map that undergoes continuous updates using real-time sensor data collected and transmitted by the robot.
Furthermore, the robot monitors wireless connectivity and automatically switches to a \emph{ autonomous mode} in scenarios with limited connectivity. 
By integrating four key functionalities: real-time mapping, remote control through glasses \gls{vr}, continuous monitoring of wireless connectivity, and autonomous navigation during limited connectivity, we achieve seamless end-to-end operation. 

\end{abstract}

\glsresetall

\section{Introduction}
In the domain of semi-autonomous robotics, addressing the significant challenge of efficient navigation and operation in environments with limited connectivity between robots and humans is imperative~\cite{Selvaggio.21}.
Creating a semi-autonomous system that provides a user-friendly experience to control robots in varied and challenging environments involves the integration of control systems, communication networks, and computation servers.
In this view, we consider a remote control setting with limited wireless connectivity where a human operator controls a robot using a \gls{vr} interface that is updated in real time from the sensory data obtained from the robot~\cite{Kalamkar.23}.
However, due to the stochastic nature of wireless connectivity, the sensing and controlling links are expected to suffer occasional losses, in which the robot’s navigation task and the corresponding real-time updates in the \gls{vr} space are likely to be interrupted.

In this work, we showcase the remote operation of a robot by a human operator in a dynamic environment. 
The operator uses a \gls{vr} interface, including a 3D map, which is updated in real-time.
If the connection between the robot and the operator is lost while navigating toward the target destination, the robot autonomously identifies the connectivity issue.
Afterward, the robot processes the sensor data onboard and adeptly navigates toward its last known destination while safely avoiding obstacles until the connection is restored.

In this view, our results highlight the adaptability of our semi-autonomous system, ensuring seamless transitions between remote control and autonomous navigation. 
Successful navigation to the last known destination during limited connectivity underscores the efficiency of our approach, promising improved human-robot interaction in dynamic environments.

\begin{figure}[!t]
\centering
\includegraphics[width=0.8\linewidth]{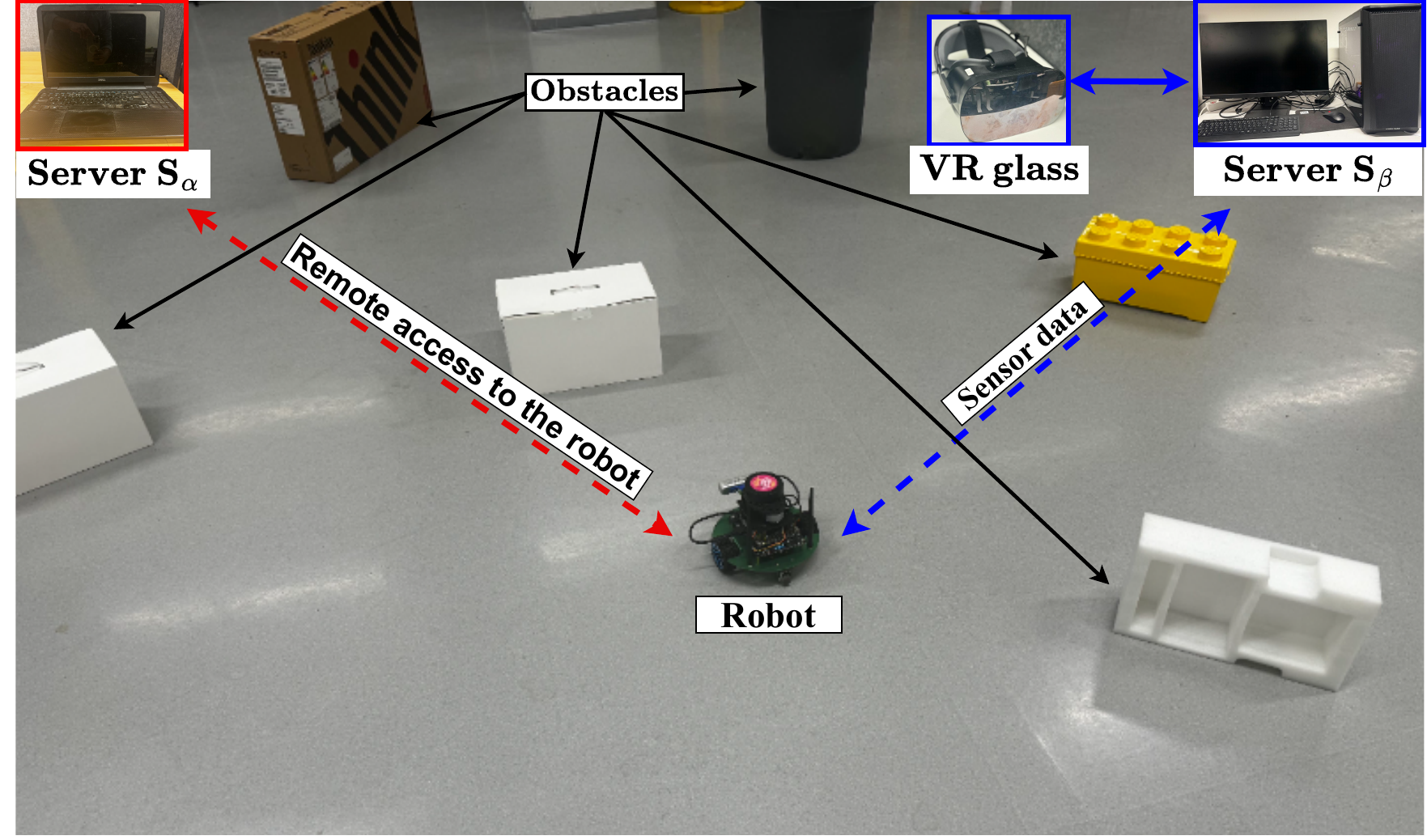}
\caption{The robotic platform highlighting the components and wireless connectivity. The red and blue lines represent the connection between each device, with dashed lines showing wireless connections and continuous lines showing wired connections.}
\label{fig:setup}
\end{figure}


\section{System Architecture \& Hardware}\label{sec:architecture}

We consider the demo setup shown in Fig.~\ref{fig:setup} to consist of a robot remotely controlled by a human using \gls{vr} glasses.  
Communication between the robot and the \gls{vr} glasses takes place through a WiFi network.
We use the Waveshare JetBot \gls{ros} \gls{ai} Kit as a mobile robot equipped with a 2D \gls{lidar} sensor, an \gls{imu} sensor, and motor encoder sensors. 
The \gls{lidar} model is RP\gls{lidar} A1, which has a scanning frequency of $5.5$\,Hz and offers a range distance of $0.15\sim12$\,m with an accuracy of $1\%$ for distances less than $3$\,m, comprising $1147$ sample points per scan~\cite{lidar}. 

In this setting, we use two servers referred to as $\server_\A$, and $\server_\B$, as shown in Fig.~\ref{fig:setup}. 
The server $\server_\A$ provides remote access to the robot and thus is used only to monitor and visualize the robot’s point of view. 
On the server $\server_\A$ we use \gls{ros} visualization (RViz), a tool built by \gls{ros}, to visualize the robot’s navigation and sensor data.
A Dell i3 laptop running on the Ubuntu \gls{os} and \gls{ros} framework is used as $\server_\A$.

The server $\server_\B$ running Windows 11 \gls{os} is used to connect the Varjo \gls{vr}-1 headset~\cite{varjo}, which acts as an interface to the human operator. 
In this way, we use Unity software to visualize the robot on a virtual 3D map with the connection facilitated through Varjo Unity \gls{xr} \gls{sdk}~\cite{unity_sdk}. 
The virtual 3D map that is processed via Unity software is used to visualize the environment through \gls{vr} glasses.
To enable communication between the robot publishing sensing data on \gls{ros} topics and the Unity software running on a Windows machine $\server_\B$, it is essential to appropriately convert the \gls{ros} topics into custom messages compatible with Unity \gls{ros}.

The robot communicates with servers $\server_\A$ and $\server_\B$ using the publish-subscribe framework of \gls{ros} for data exchange.
The data published includes information about obstacle locations, path planning information, live video feed, and the robot's odometer readings.
Data are sent to the remote server $\server_\B$ in a custom \gls{ros} message format. 
To facilitate data transfer, we utilize the Unity \gls{ros} \gls{tcp} connector
on server $\server_\B$.

\label{sec:slam_map}
\begin{figure}[!t]
\centering
\includegraphics[width=0.8\linewidth]{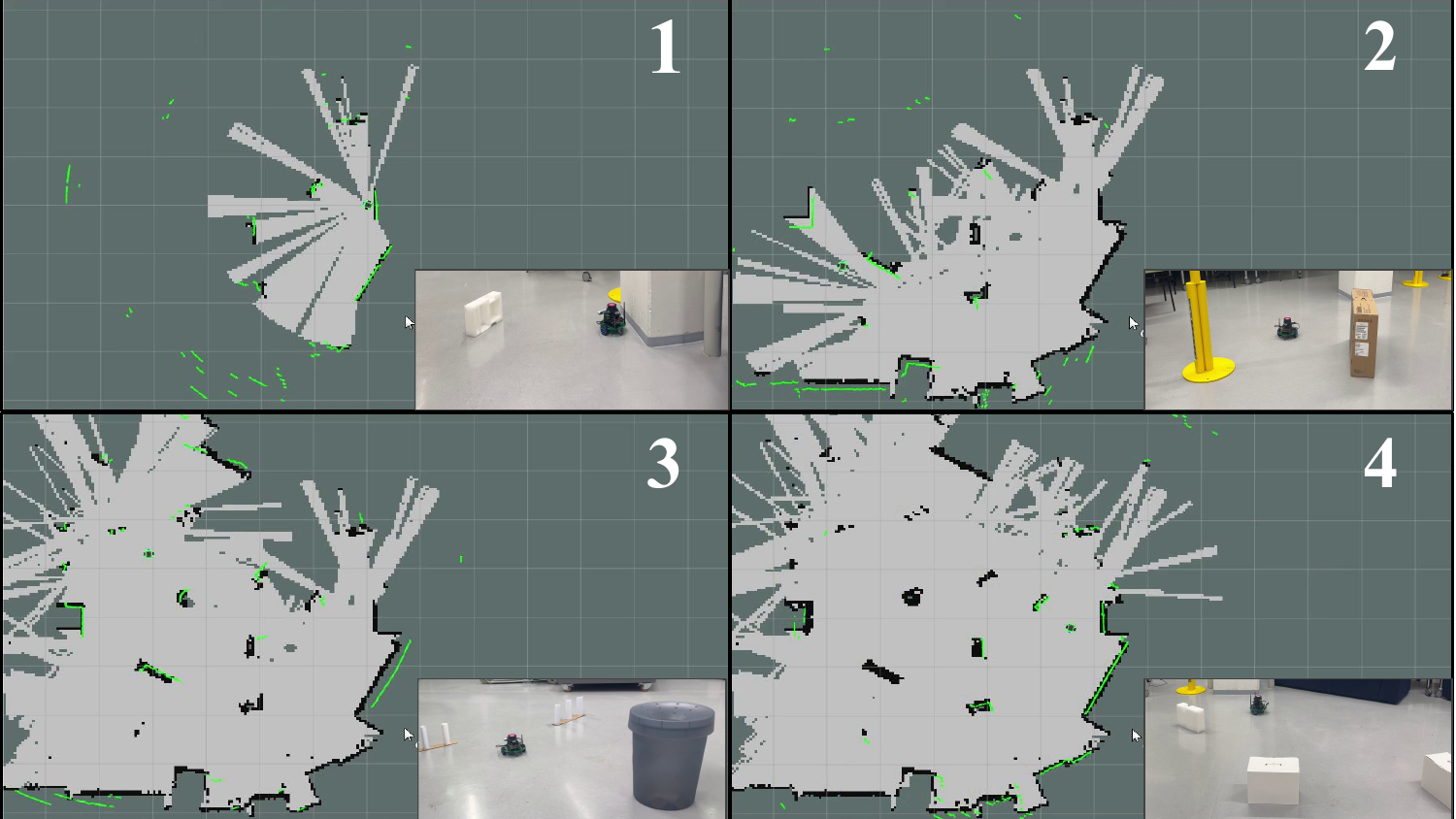}
\caption{The visualization of SLAM showing different areas: collision-free (light gray), unexplored (dark green), occupied and inaccessible areas (black).}
\label{fig:slam_map}
\end{figure}

\section{System Implementation}

This demo is composed of four key software solutions: (i) creating a 2D map for navigation, (ii) remote control through \gls{vr} glasses, (iii) detection of the status of robot-sever connectivity, and (iv) autonomous navigation. 
The following sections provide detailed information on each of these aspects.

\subsection{Creating a 2D Map for Navigation}

Our demo project showcases scenarios where a robot is remotely controlled under occasional changes in the destination set by a human operator in unknown areas.
To facilitate a better experience for the operator as well as to be able to autonomously navigate under possible losses in remote connectivity, it is essential to generate a map of its operating environment.
To this end, we implement a \gls{slam} technique that can adapt to dynamic changes in the environment.
\Gls{slam} plays a central role in the field of robotics and autonomous systems. It enables autonomous vehicles to create maps of their surroundings while simultaneously determining their own position within these maps.

Here, we utilize the TurtleBot3 \gls{slam} gmapping package \gls{ros}~\cite{turtlebot3} to create the 2D environment map.
As illustrated in Fig.~\ref{fig:setup}, our test area contains several obstacles.
Then, as the robot moves around this test area, \gls{slam} uses the 2D \gls{lidar} data to create the environment map.
We can visualize the creation of the 2D map using $\server_\A$, as shown in Fig.~\ref{fig:slam_map}. 
In $\server_\A$ we can observe the different areas of the 2D maps, such as collision-free, unexplored, occupied, and inaccessible areas. 
Fig.~\ref{fig:slam_map} shows the real-time discovery and construction of the 2D map with each area represented by a different color.
The map is constructed as a 2D \gls{ogm}, commonly used in the \gls{ros} community.
After exploring the test area, the robot saves the \gls{ogm} map locally for future use.

\subsection{Remote control via VR glasses} 
\label{sec:manual_mode}

\begin{figure}[!t]
\centering
\includegraphics[width=\linewidth]{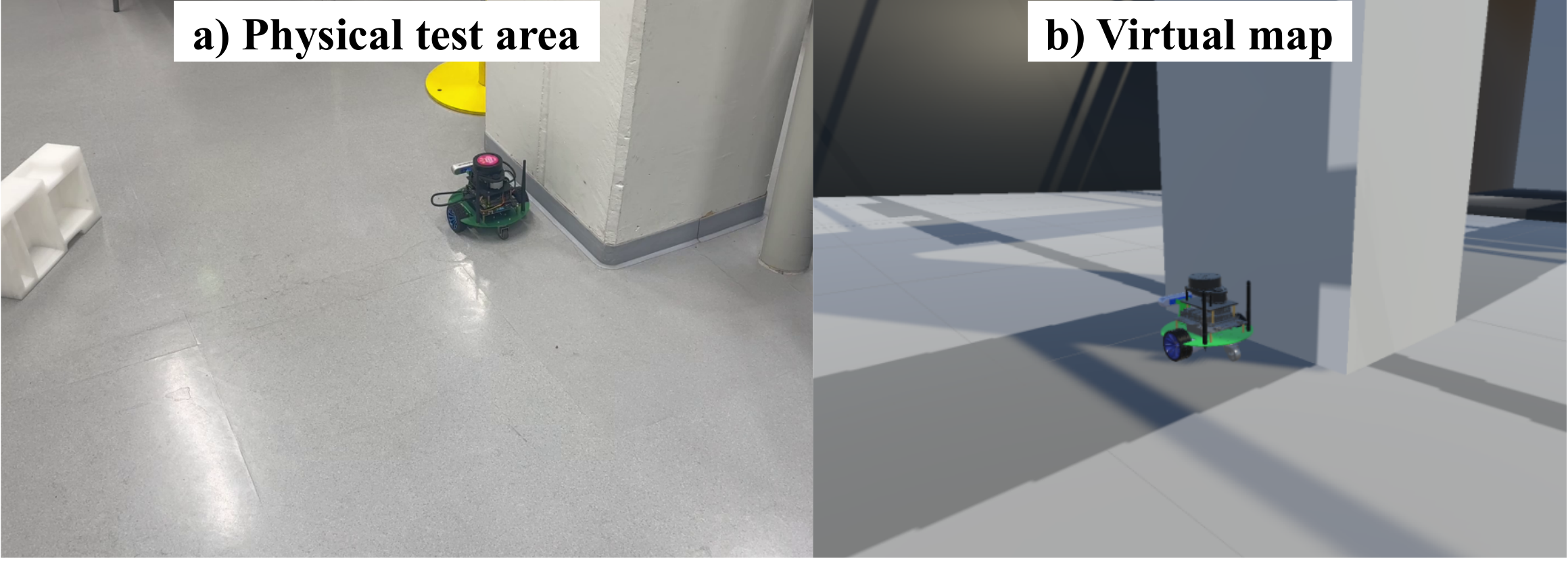}
\caption{Initial position of the robot on \textit{a)} the physical test area and \textit{b)} the virtual map.} 
\label{fig: initial pose}
\end{figure}

In this section, we explore remotely maneuvering the robot to a specified destination using \gls{vr} glass, referred to as the \emph{remote control mode}. 
We use a 3D virtual map of our test area~\cite{virtual_map} and set the robot's initial position approximately the same on both the virtual map and the physical test area, as illustrated in Fig.~\ref{fig: initial pose}.
Afterward, the operator sets a destination on the previously created 2D map.

\begin{figure}[t]
\centering
\includegraphics[width=0.75\linewidth]{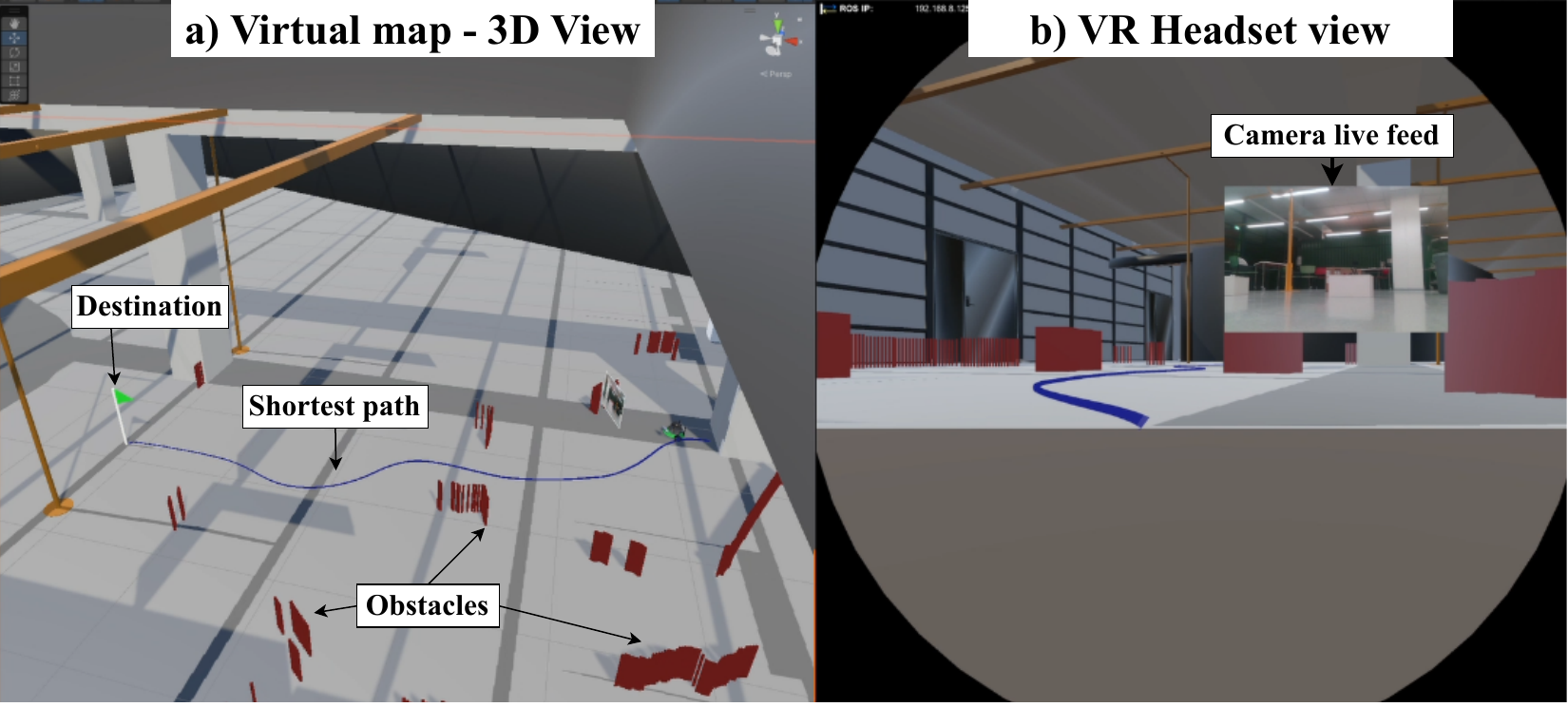}
\caption{The 3D virtual environment at the operator's end.} 
\label{fig:vr_view}
\end{figure}

In the 3D virtual map, the position and orientation of the virtual robot are continuously updated using the odometry data received from the physical robot.
Meanwhile, the \gls{lidar} data offers insight into obstacles within the test area to $\server_\B$.
Although $\server_\B$ does not have object recognition capability, it is crucial to provide obstacle information through the \gls{vr} interface to the human operator for remote control. 
In this view, obstacles are visualized as brown wall segments in the virtual map, which are updated in real-time on both the 3D map and on the \gls{vr} headset view, as illustrated in Figs.~\ref{fig:vr_view}a and ~\ref{fig:vr_view}b, respectively.
To validate the accuracy of the real-time visualization, the robot's live video feed is also projected on the \gls{vr} view, as shown in Fig.~\ref{fig:vr_view}b.

It is worth highlighting that real-time remote control and virtual map updating are based on the wireless connectivity between the robot and $\server_\B$. 
During a loss of connectivity, both sensing and controlling fail, and thus, the navigation procedure gets interrupted. 
To ensure seamless operation, it is essential to devise mechanisms to detect connectivity loss and to be able to switch from remote control to autonomous mode (and vice versa).

\subsection{Detection of the robot-server connectivity status}
\label{sec:identifying_Limited_connectivity}

Here, we discuss the process of identifying the loss of connectivity between the robot and the server $\server_\B$. 
To identify the quality of the connection between the robot and the server $\server_\B$, the robot periodically sends a \texttt{ping} command with an interval of $0.1$s directly to a known \gls{ip} address of $\server_\B$. 
If the response code from server $\server_\B$ is $0$, it indicates a good connectivity status, otherwise it suggests poor or no connectivity.
As long as the connection status is good, the robot operates in the remote control mode and can be operated by the human operator. 
In any other case, the robot switches from remote control mode to \emph{autonomous mode}, which is discussed next. 

\subsection{Autonomous navigation}
\label{sec:Autonomous_mode}

The \emph{autonomous mode} allows the robot to operate seamlessly in the absence of connectivity to the remote operator.
Here, navigation is based on computing the shortest path from the current position to the last known target location using the TurtleBot3 Navigation \gls{ros} package~\cite{turtlebot3}. 
Upon detecting a loss of connection between the robot and the $\server_\B$, the robot publishes the target location data using
\texttt{geometry\_msgs/PoseStamped} \gls{ros} message type from the \texttt{/move\_base\_simple/goal} rostopic 
to enable the functionality of autonomous navigation.

\begin{figure}[t]
\centering
\includegraphics[width=0.8\linewidth]{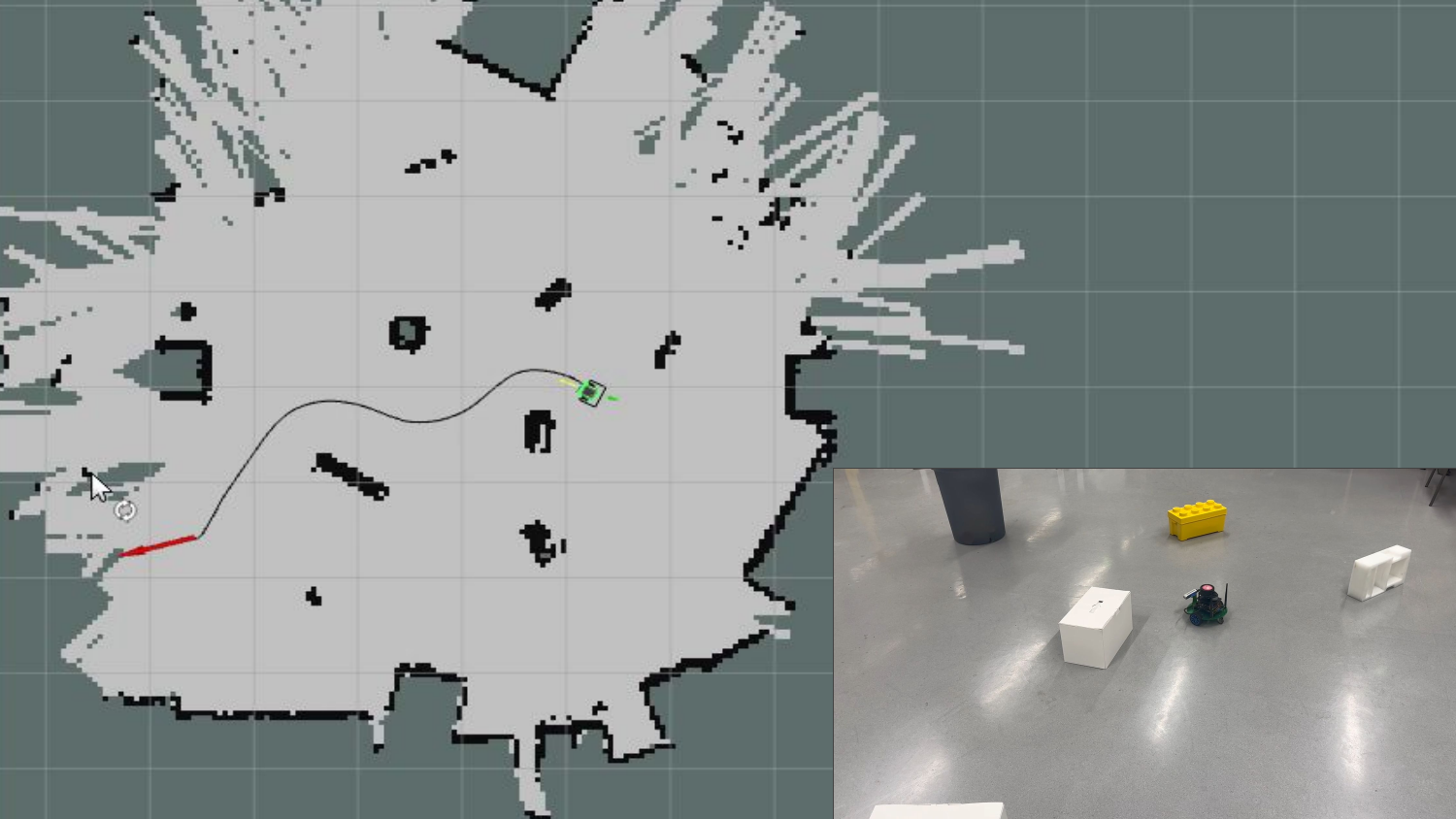}
\caption{A snapshot at the $\server_\A$ during the autonomous mode. Since there is no remote connectivity and control, the robot (black square) follows the calculated trajectory (black curve) towards the last known destination (red arrow).}
\label{fig:auto_rviz_view}
\end{figure}

It is worth highlighting that any real-time changes taking place in the environment are perceived by the \gls{lidar} sensors, and therefore the \gls{ogm} and the cost map are updated simultaneously. 
The cost map is a grid in which every cell is assigned a value (cost) that determines the distance from the obstacle, where a higher value means a closer distance.
The cost map is provided by the TurtleBot3 Navigation package \gls{ros}.
With \gls{ogm} and the dynamic information from the cost map, the robot continuously refines its planned trajectory, focusing on the shortest collision-free path, ensuring collision-free navigation.
Fig.~\ref{fig:auto_rviz_view} shows a snapshot on the server $S_\A$ during autonomous navigation mode, illustrating the calculated shortest path that the robot follows.
Since the robot simultaneously checks the connectivity status, upon restoring the connection with $\server_\B$, the operation mode is switched back to \emph{remote control} and the procedure defined in Sec.~\ref{sec:manual_mode} is carried out with the modifications made by the human operator.

During the \emph{autonomous mode}, no sensing data are available at $\server_\B$, yet the virtual robot on the virtual map is designed to follow the last computed shortest path at a constant speed to mimic the behavior of the physical robot.
Displaying the last computed shortest path through \gls{vr} glasses serves two purposes in enhancing the seamless experience for the human operator:
(i) The continuous updates in the virtual world are used to avoid lags and freezes in the \gls{vr} interface 
and
(ii) the risk and the degree of the virtual robot's teleportation (sudden jump from one location to another) upon the connection restoration is reduced.

\subsection{Demo, Resources, and Future Developments}

The software related to the robot and servers is available at 
\url{https://github.com/ICONgroupCWC/Real-Time_VR_RemoteControl}. Future directions include the fusion of 3D \gls{lidar} point cloud data and depth camera information and the development/improvement of 3D \gls{slam} methods to improve accuracy and understanding of the environment for real-time robotic navigation.
This demo in action can be seen from \url{https://youtu.be/1Hd78-bGPe0} 
%

\bibliographystyle{IEEEtran}
\bibliography{reference}

\end{document}